\documentclass[letterpaper, 10 pt, conference]{ieeeconf}
\IEEEoverridecommandlockouts
\usepackage{amsmath,amssymb}
\usepackage{leftidx}
\usepackage{xcolor}
\usepackage{hyperref}
\usepackage{cleveref}
\usepackage[T1]{fontenc}
\usepackage[utf8]{inputenc}

\usepackage[noadjust]{cite}
\usepackage[pdftex]{graphicx}
\graphicspath{{images/}}
\DeclareGraphicsExtensions{.pdf,.jpeg,.png}
\usepackage[caption=false,font=footnotesize]{subfig}

\title{\LARGE \bf 3D mapping for multi hybrid robot cooperation}

\author{Hartmut Surmann$^{1}$ and Nils Berninger$^{1}$ and Rainer Worst$^{2}$
\thanks{TRADR is funded by EU-FP7-ICT grant No. 609763. TRADR website: http://www.tradr-project.eu/.}
  \thanks{$^{1}$University of Applied Science Gelsenkirchen, Fraunhofer Institute for Intelligent Analysis and Information Systems IAIS, Schloss Birlinghoven  53757 Sankt Augustin, Germany {\tt\small hartmut.surmann@w-hs.de}}
  \thanks{$^{2}$Fraunhofer Institute for Intelligent Analysis and Information Systems IAIS, Schloss Birlinghoven 53757 Sankt Augustin, Germany
  {\tt\small rainer.worst@iais.fraunhofer.de}}
}

\begin{document}

\maketitle
\thispagestyle{empty}
\pagestyle{empty}

\begin{abstract}
 This paper presents a novel approach to build consistent 3D maps for multi robot cooperation in USAR environments.
 The sensor streams from unmanned aerial vehicles (UAVs) and ground robots (UGV) are fused in one consistent map.
 The UAV camera data are used to generate 3D point clouds that are fused with the 3D point clouds generated
 by a rolling 2D laser scanner at the UGV. The registration method is based on the matching of corresponding planar segments
 that are extracted from the point clouds.
 Based on the registration, an approach for a globally optimized localization is presented.
 Apart from the structural information of the point clouds, it is important to mention that no further information is required for the localization.
 Two examples show the performance of the overall registration.
\end{abstract}

\section{Introduction}

\begin{figure}[!t]
\centering
\subfloat[UAVs]{\includegraphics[width=0.54\linewidth]{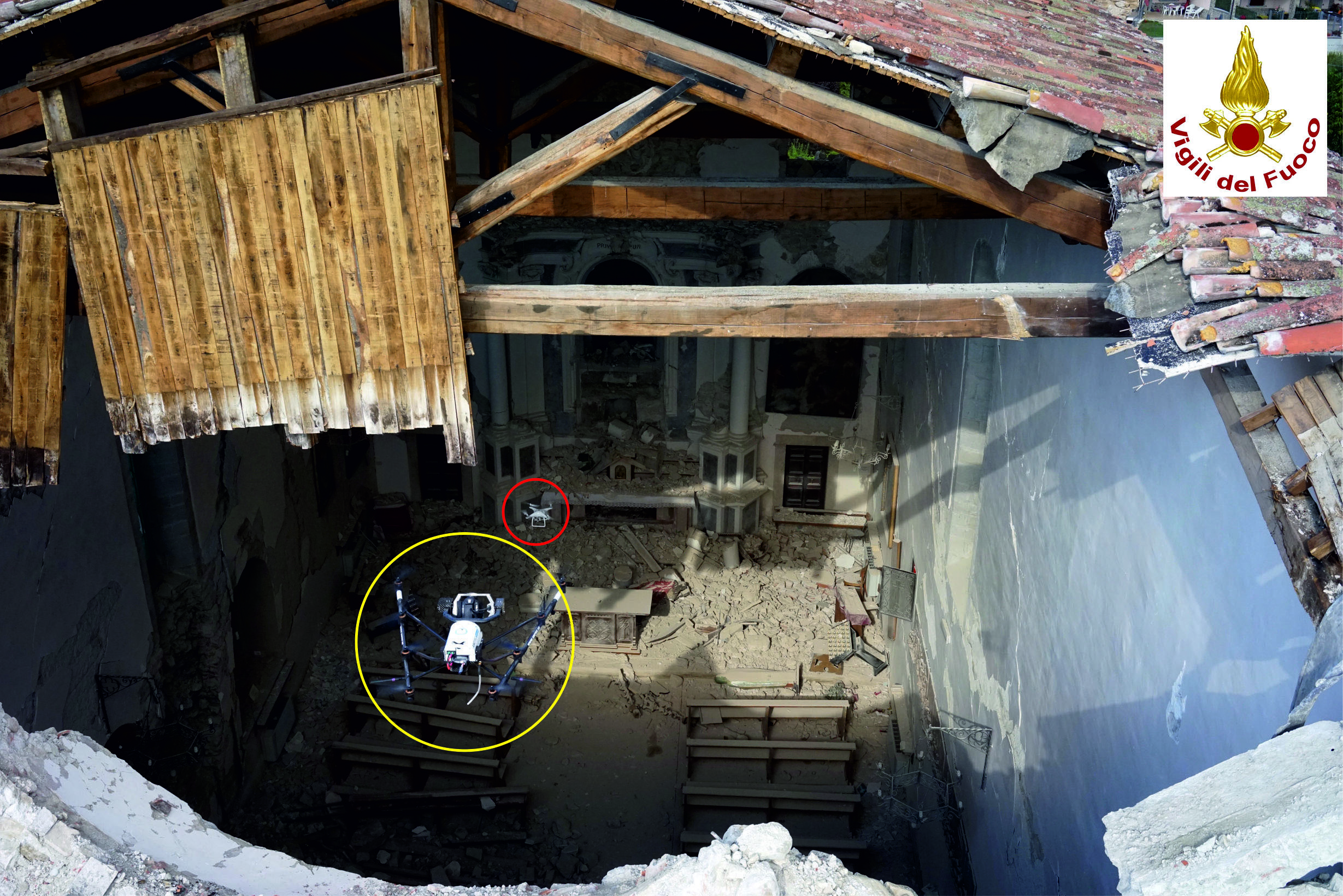}}
\hfil
\subfloat[UGV]{\includegraphics[width=0.44\linewidth]{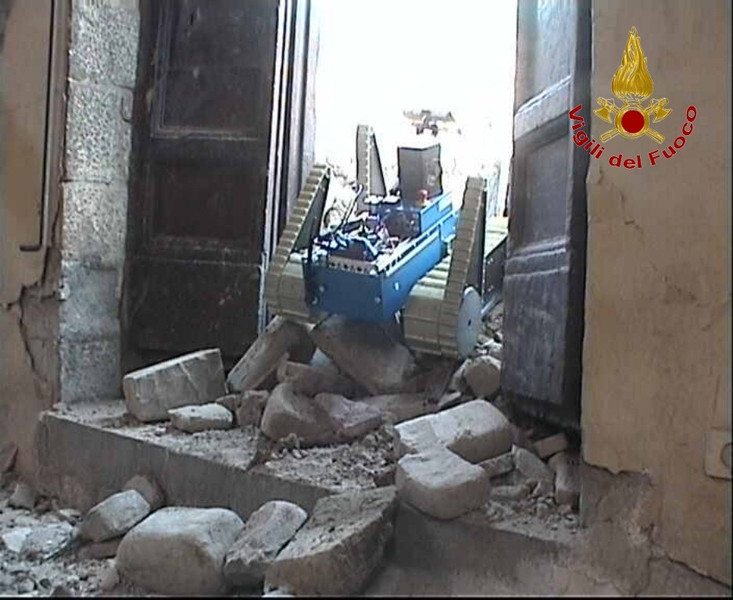}}
\caption{UAVs and UGV of the TRADR project after the earthquake in Amatrice / Italy 2016.}
\label{fig:amatrice}
\end{figure}

Despite the growing technological advances, coping with disaster scenarios is still a major challenge for robots and humans.
After 48 hours the probability of rescuing people from a collapsed building is drastically reduced \cite{murphy2000marsupial}.
The EU project TRADR develops novel science \& technology for human-robot teams to assist in disaster
response efforts, over multiple sorties during a mission. Various kinds
of robots collaborate with human team members to explore the environment and
to gather physical samples (fig. \ref{fig:amatrice}).
The goal is to enable the team to gradually develop its understanding of the disaster area over multiple, possibly
asynchronous sorties (persistent environment models), to improve team members'
understanding of how to work in the area and to improve team-work.
The fusion of different sensor streams of different semi-autonomous robots (UAV and UGV) in one consistent map is the basis for building the environment models.
The UGV can be equipped with several sensors, e.g. tilting laser scanners and even actuators due to its higher payload. The UAV is equipped with only a few light sensors to reduce the weight and to increase the flight time.
According to the current state of the UAV market, it is only possible to obtain important information in real-time by using a monocular camera.
Furthermore, additional sensors such as GPS are available for localization, but a reliable position determination cannot always be ensured depending on the environment, e.g. close to buildings.
The nature of the resulting data, which differ due to different sensors, is a major challenge.
In order to use the data collaboratively, representations and algorithms have to be found that can process data from different sources (more or less in real-time).

For this work, a UGV with a laser scanner and a UAV with a monocular camera are used. While a laser scanner can directly provide distance information, more complex methods -- generally known as Structure from Motion (SfM) -- are used for extracting distance information from camera recordings. The difficulty when combining the data is finding corresponding regions that allow a robust transformation between the two point clouds.
Naively, this should be possible with standard point-based scan matching. Unfortunately, point-based ICP fails due to the differences of the point clouds from the different sensors. Laser scanners compute precise radial point clouds whereas SfM or multi-view stereo algorithms compute less precise, erroneous, and non equally distributed dense point clouds, focused on brightness differences and textures.      
Therefore, geometrical structures have to be described as invariant as possible against the individual disturbances of the different sensors.

The objective of this work is the development of a method for a typical rescue scenario. The first responder arrives at the disaster site and uses the UAV to get an overview, i.e. images and an initial 3D point cloud. Humans and the UGVs use this initial sensor streams, which allow the localization of the UGV in a vision-based map.
The approach uses surfaces that abstract from the underlying data structure and hence can compensate disturbances while still containing sufficient information for the motion estimation.
The resulting 3D map combines the information from both sensors and thus has a higher information content.
The collected data of the UAV have to be processed by a SfM method independent of the UGV. Subsequently, the results of the processing can be provided to the UGV for a first localization.

The remaining paper is organized as follows: The next section summarizes state of the art methods that can be used to generate point clouds from camera recordings. Various registration methods are also reviewed and \cref{sec:pose_tracking} presents the selected registration method. An example of how we used this method for globally optimized localization is given in \cref{sec:localization}. Several results of our experiments are shown in \cref{sec:experiments}. Videos can be found at youtube {www.youtube.com/watch?v=xAVR5aFv8VY.

\section{Related work}
\label{sec:related_work}
A basic prerequisite for many tasks, such as navigation, mapping or cooperation of UAV and UGV, is the robot localization. When working with three-dimensional point clouds, the registration is significantly affected by the success of an exact localization \cite{holz2015registration}. Due to the aim of this work, to localize the UAV and UGV together in a global map, a registration method has to be found that can handle point clouds from different sources. In this context, it is important that the methods for registration as well as the generation of vision-based point clouds can be combined.

\subsection{Vision-based SLAM}
\label{subsec:vision_based_slam}
In order to perform visual odometry, only keypoints are selected, which make a robust correspondence search possible. While some methods compute complex features (\cite{klein07parallel, davison2007monoslam}), new developments increasingly use image points directly (\cite{newcombe2011dtam, engel2013iccv, Forster2014ICRA}). Direct approaches have the advantage that they are not reduced to certain feature points but can exploit all image points to determine the odometry and depth values and thus provide more dense reconstructions of the environment. Depending on how many image points are utilized, the approaches can be divided into dense and semi-dense methods.

An example of a semi-dense approach is the SVO algorithm, which is presented in the work of \cite{Forster2014ICRA}. The method uses point features, but these are not explicitly extracted. Rather they are an implicit result of a direct motion estimation. The initialization of the pose is achieved by minimizing the photometric error. LSD-SLAM \cite{engel14eccv} provides another direct approach. Based on the odometry method of \cite{engel2013iccv}, the algorithm generates globally consistent maps of the environment by means of graph optimization in large-area environments. Similar to the SVO algorithm, a probabilistic representation of the depth map is also used here to model inaccuracies. \cite{mur2015probabilistic} also uses a probabilistic approach, but the method is based on a feature-based monocular SLAM system (\cite{murAcceptedTRO2015}). Furthermore, in contrast to SVO and LSD-SLAM, the depth values of a reference image are not filtered over many individual images, but only key images are used for the reconstruction.

\cite{stuehmer-et-al-dagm10} presents one of the first real-time methods and provides dense reconstructions with a monocular camera. The tracking of the camera is based on the approach of \cite{klein07parallel}. The reconstruction is carried out using several key images. By expanding to several images, regions that would be hidden in two images or would be outside the corresponding image can also be reconstructed with a higher probability. DTAM (\cite{newcombe2011dtam}) also provides dense reconstructions in real-time. In order to estimate the depth values, the method performs a global energy reduction over many individual images. REMODE (\cite{Pizzoli2014ICRA}) is a method for the reconstruction of dense point clouds, which integrates a Bayesian estimate into the optimization process. By modeling uncertainties of measurement for each pixel, regularization can be controlled precisely and inaccuracies in the localization can be reduced. Real-time capability is achieved through a CUDA-based implementation. For the pose estimation, the method of \cite{Forster2014ICRA} is used. One of the recent developments of dense reconstructions is DPPTAM \cite{conchaIROS15}. The approach reconstructs high textured regions with a semi-dense approach and low textured regions by approximation of surfaces. Thereby the assumption is made that homogeneously colored image regions form a plane, which can be determined by superpixels (\cite{felzenszwalb2004efficient}).

The procedures described so far fall under the category of online procedures, i.e. they are real-time capable and can deliver first results during camera recording. In contrast, offline procedures require all collected recordings in advance and then carry out the corresponding calculations. In \cite{fuhrmann2014mve}, a pipeline for reconstruction is presented that combines all necessary processing steps in a software framework called MVE. The framework is also capable of reconstructing texturized surfaces.

\subsection{Registration methods}

Methods for registration can be divided roughly into point-based or iterative and feature-based methods (\cite{holz2015registration, Pathak2009}). An example of a known iterative method is the ICP-algorithm, which has already been implemented in several variants. According to \cite{Besl1992} the transformation is determined by minimizing the Euclidean distance of the found point correspondences. The search for corresponding points and the calculation of the associated transformation for the alignment of these points is finally repeated iteratively until pre-defined limits have been reached. A disadvantage of iterative methods, however, is that they can converge to a local minimum under certain assumptions, such as an insufficient overlay of the scenes \cite{holz2015registration}. In addition, they can be sensitive to outliers and can be very computationally intensive with large amounts of data \cite{viejo20073d}. If several point clouds have to be registered, the generated scene must also be globally consistent. To achieve better results, it is common that feature-based methods are used for the initial registration and iterative methods are used for refining the already estimated transformation \cite{holz2015registration}. Features can be described by feature descriptors that incorporate geometric structures. If surfaces are used as a geometric structure, a high compression rate and thus a fast correspondence search can be achieved \cite{Pathak2009}. 

The work of \cite{PathakJFR2010} introduces a SLAM algorithm based on the registration of planar segments. The algorithm for the extraction of planar segments is based on the work of \cite{poppinga2008fast}, which takes up the region-growing algorithm of \cite{hahnel2003learning} and adapts it by optimizations for the use in a SLAM system. For correspondence search and registration, the work of \cite{pathak2010fast} is used. The presented MUMC-algorithm (Minimally Uncertain Maximum Consensus) maximizes geometric consistency while minimizing the resulting uncertainties. As shown in the work of \cite{PathakJFR2010}, both faster and more robust results can be obtained in comparison to an ICP-alorithm. \cite{xiao2013planar} provides another plane-based registration method, which is based on the work of \cite{PathakJFR2010}. An approach that is also concerned with the registration of point clouds from different sensor groups is presented in \cite{gawel2016structure}. As a first step, the method determines structural descriptors. For faster calculation, the descriptors are then projected into a subspace. A matching scheme is used to compare the descriptors and compute vote scores. The voting space is then used for place segmentation and for registration.

For this work, an algorithm is developed that is based on the approaches of \cite{xiao2013planar} and \cite{PathakJFR2010}. The presented algorithm for surface extraction can be applied to unorganized point clouds and is fast in the calculation. The method of \cite{PathakJFR2010} has also proven itself in a test environment that is very close to a possible application area of this work.

\section{Pose tracking}
\label{sec:pose_tracking}

This section introduces the registration method used for relative localization.
The first step is the segmention of planes from the source and the target point cloud as described in \cite{xiao2013planar}. Afterwards corresponding planes and the associated transformation must be determined. The correspondence search is based on \cite{xiao2013planar}, but in contrast to the original algorithm, the area of the planes is determined by \cite{Marton09ICRA}. In addition, the correspondence search was extended by the examination of overlapping planes. This is done as follows: First, the transformation determined on the basis of corresponding planes is temporarily applied to the planes to be examined. Then the minimum and maximum coordinate values of each plane are determined and the vectors $v_{min}$ and $v_{max}$ are formed. Two planes $^{d}P$ and $^{m}P$ are overlapping when

\begin{equation}
\leftidx{^m}{v_{min}} < \leftidx{^d}{v_{max}} + \varepsilon \quad \text{and} \quad \leftidx{^d}{v_{min}} < \leftidx{^m}{v_{max}} + \varepsilon
\end{equation}

\noindent
is satisfied. Here $\varepsilon$ is a positive number that defines a tolerance range. The directions along the surface normals of the target planes can be ignored during verification.

The last step is to determine an optimal transformation from all corresponding planes as described in \cite{PathakJFR2010}. The rotation and translation is calculated in a separate step. A plane $P$ will be defined by its oriented and normalized surface normal $\hat{n}$ and the distance $d$ to the coordinate origin. If the correspondence set $\Omega = \{\langle \leftidx{^m}{P_{i1}}, \leftidx{^d}{P_{i2}} \rangle, \; i = 1, \ldots, N_\Omega\}$, which assigns every plane $\leftidx{^d}{P_{i2}}$ of the source point cloud a corresponding plane $\leftidx{^m}{P_{i1}}$ of the target point cloud, is known, the optimal rotation can be calculated by minimizing the following function:

\begin{equation}
f\left(R\right) = \cfrac{1}{2} \sum_{i = 1}^{N_{\Omega}} ||R\,\leftidx{^d}{\hat{n}_{i2}} - \leftidx{^m}{\hat{n}_{i1}}||^2.
\end{equation}

\noindent
The translation is expressed by the equation

\begin{equation}
N \leftidx{^m_d}t = d,
\end{equation}
with
\begin{equation}
\quad N_{N_{\Omega} \times 3} =
\begin{bmatrix}
\leftidx{^m}{\hat{n}_{1}}^T\\
\vdots\\
\leftidx{^m}{\hat{n}_{N_{\Omega}}}^T
\end{bmatrix} \quad \text{and} \quad d_{N_{\Omega} \times 1} =
\begin{bmatrix}
\leftidx{^m}{d_{1}} - \leftidx{^d}{d_{1}}\\
\vdots\\
\leftidx{^m}{d_{N_{\Omega}}} - \leftidx{^d}{d_{N_{\Omega}}}
\end{bmatrix}.
\end{equation}

\noindent
The equation is solved by means of singular value decomposition. The singular value decomposition of the matrix $N$ is given by

\begin{equation}
N_{N_{\Omega} \times 3} = U_{N_{\Omega} \times N_{\Omega} } \Sigma_{N_{\Omega}  \times 3} V_{3 \times 3}^T.
\end{equation}

\noindent
Here the column vectors $u_i$ of $U$ are the left singular vectors and the column vectors $v_i$ of $V$ are the right singular vectors.

$\Sigma$ is a $N_{\Omega} \times 3$ diagonal matrix, which contains the real positive singular values $\sigma_i$. Afterwards, a rank decision for the matrix $N$ will be made, i.e. the rank $r$ will be chosen so that $\sigma_r > 0$ and $\sigma_{r+1} = \dots = \sigma_n = 0$.

The best approximation of  $N$ is given by $\hat{N}_r$ with

\begin{equation}
\hat{N}_r = \sum^{r}_{i = 1} \sigma_i u_i v_i^T \text{,}
\end{equation}

\noindent
The best translation estimation for rank $r$ can finally be achieved by

\begin{equation}
\label{eq:opt_translation_1}
\leftidx{^m_d}t = \sum^r_{i = 1} \sigma_1^{-1} \left(u_i \cdot d\right) v_i.
\end{equation}

\noindent
If two laser point clouds are compared with one another, the ICP algorithm can be used for further refining the translation. The transformation already determined serves as an initial position estimation. Another option is given by the odometry estimation of the robot. If an additional translation estimation $\leftidx{^m_d}t_e$ could be computed, it can be used to determine the missing translation directions $v_i,\, i = r + 1,\ldots,3$ and can be integrated with

\begin{equation}
\leftidx{^m_d}t = \sum^r_{i = 1} \sigma_1^{-1} \left(u_i \cdot d\right) v_i + \sum^3_{i = r + 1} \sigma_1^{-1} \left( \leftidx{^m_d}t_e \cdot v_i \right) v_i
\end{equation}

\noindent
in the overall translation estimation.

\section{Localization} 
\label{sec:localization}
This section describes how the registration procedure described in \cref{sec:pose_tracking} can be used for localization and mapping.
In robot localization, a distinction can be made between relative and absolute localization. In the case of relative localization, the changes in the respective current pose are determined from a known pose and thus the entire trajectory is built step by step. In the absolute localization, the pose is determined with respect to a given map. A disadvantage of the relative localization is that errors in the determination of the pose changes are accumulated and thus the estimated trajectory as well as the constructed map are not globally consistent. However, if the starting position within a given map is known, the relative localization can be optimized. For each update step, the new pose is compared with the given map and a correction is made. The global map is provided by the UAV, which takes images during a first flight over the environment and generates a point cloud by means of a vision-based SLAM algorithm. If the absolute pose of the UGV is known in the global map, the map can be extended by the information of the laser scan and a more detailed map can be built step by step. This is useful, on the one hand, in low-textured regions, which cannot be covered by most camera-based methods. On the other hand, map areas such as interiors that are not accessible to the UAV or that are not visible in the event of a flyover due to occlusions can also be included in the global map. All processing steps involved are explained below; see \cref{fig:registration_pipeline} for an overview of the whole process.

\begin{figure}[htbp]
\centering
\includegraphics[width=0.49\textwidth]{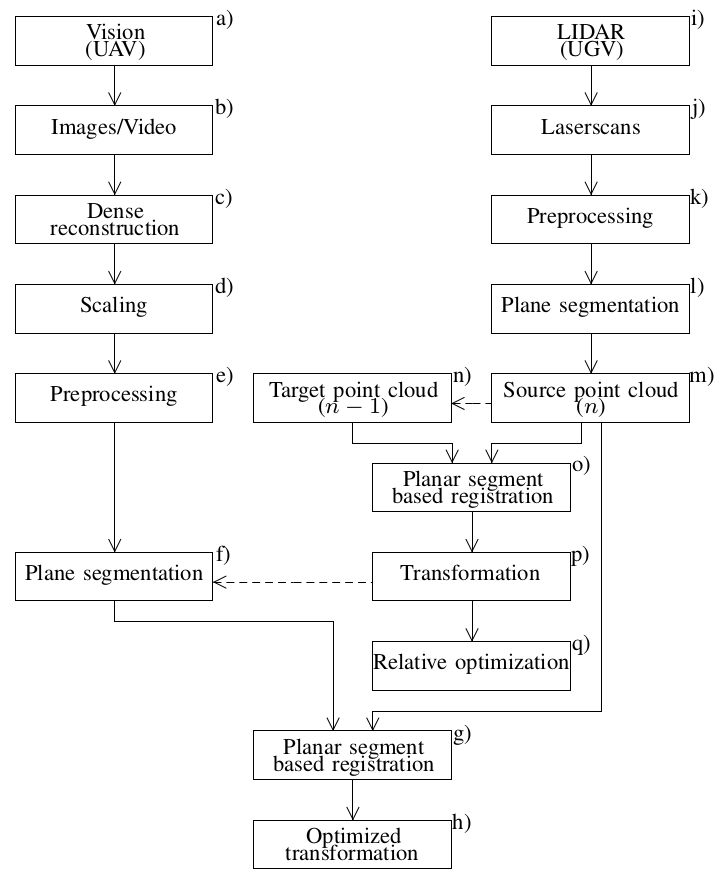}
\caption{Localization pipeline for laser point clouds with following global optimization step.}
\label{fig:registration_pipeline}
\end{figure}

Steps a--c of \cref{fig:registration_pipeline} describe the dense reconstruction with a suitable algorithm, e.g. MVE. After a reconstruction, the vision-based point clouds have to be scaled (\cref{fig:registration_pipeline}, d). This is necessary since the scaling factor for the reconstruction cannot be unambiguously determined when using a monocular camera. A correct scaling factor can be determined in several ways. For this work, the GPS coordinates recorded by the UAV during its flight are used. For the calculation, all positions estimated using the vision-based method are assigned to the nearest GPS coordinates and the Euclidean distances of adjacent points are calculated. The ratio of the average distances finally indicates the scaling factor.

As preparation for the plane segmentation, the point clouds are filtered through several processing steps (\cref{fig:registration_pipeline}, e and k). The aim of the preprocessing is to increase the robustness of the plane segmentation and thus the subsequent registration. By filtering, the point clouds are also reduced in size, which can considerably reduce the computational effort. For this work, a voxel grid filter and an outlier removal filter are applied, but additional filters can be added if necessary.

The relative pose of the UGV is updated with each new laser scan (\cref{fig:registration_pipeline}, l--p). First of all, a plane segmentation is carried out once for each new point cloud. Then the relative transformation between the last and the last but one point cloud is determined by means of a planar segment-based registration. For the initial laser point cloud, the assumption is made that it has approximately the correct pose with respect to the global map of the UAV. One way to determine the pose is by matching GPS coordinates. An exact pose is not necessary, since the initial pose is subsequently adjusted as part of the global optimization. If the initial pose was determined, an accumulation of the relative transformations can be used to estimate the current global pose. This pose will be optimized by aligning the associated point cloud with the global point cloud of the UAV. To achieve this, a planar segment-based registration is performed between the current laser point cloud and a section of the global vision-based point cloud (\cref{fig:registration_pipeline}, f--h). The position and size of the section is determined by the position and size of the current laser point cloud. Since the relative transformation is not always exact, the section is additionally expanded by a tolerance range. If the UGV moves in regions that are not or only slightly captured in the global point cloud of the UAV, a global optimization is not possible. In this case, a relative optimization can be carried out using a metascan algorithm (\cref{fig:registration_pipeline}, q). For this purpose the \textit{simultaneous matching} algorithm from \cite{surmann2003autonomous} was adapted for a surface-based approach and used for this work as follows:

\begin{enumerate}
\item 
The first point cloud that could no longer be optimized globally is defined as the master point cloud and determines the coordinate system. The already calculated relative transformation of a new point cloud serves as the initial registration of the relative optimization.
\item A list is initialized with the new point cloud.
\item The following three steps are repeated until the list contains no more elements:
\begin{enumerate}
\item The first point cloud in the list is removed as the current point cloud from the list.
\item If the current point cloud is not the master point cloud, then the neighbouring point clouds of the current point cloud are calculated. 
A point cloud is regarded as a neighbouring point cloud when a given minimum number of surfaces overlap with the surfaces of the current point cloud. 
All neighbouring point clouds are then grouped into a single point cloud and a planar segment-based registration with the current point cloud is performed.
\item If the calculated transformation changes the pose of the current point cloud by more than a pre-defined minimum, all neighbouring point clouds that are not already in the list are added to the list.
\end{enumerate}
\end{enumerate}

\Cref{fig:metascan} illustrates a possible result of the relative optimization.

\begin{figure}[]
\centering
\subfloat[Point cloud 1]{\includegraphics[width=0.32\linewidth]{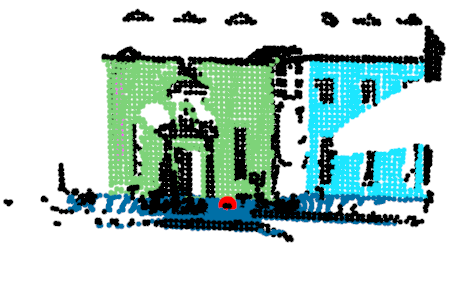}}
\hfil
\subfloat[Point cloud 2]{\includegraphics[width=0.32\linewidth]{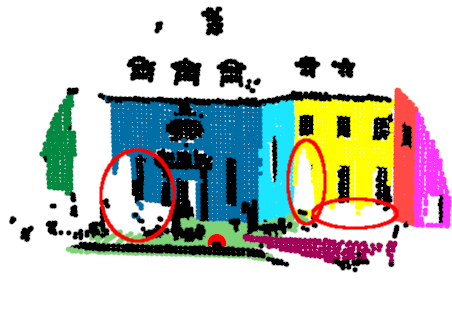}}
\hfil
\subfloat[Point cloud 3]{\includegraphics[width=0.32\linewidth]{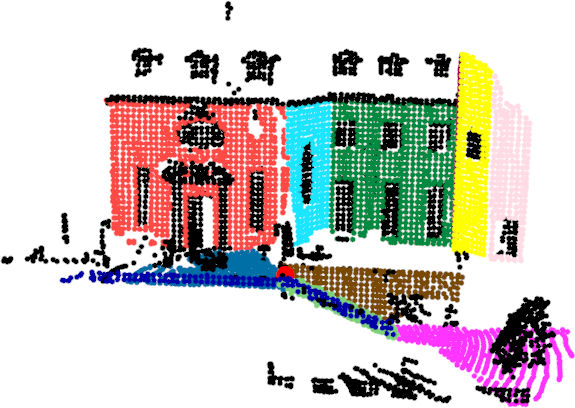}}
\hfil
\subfloat[Point clouds 1 and 2]{\includegraphics[width=0.49\linewidth]{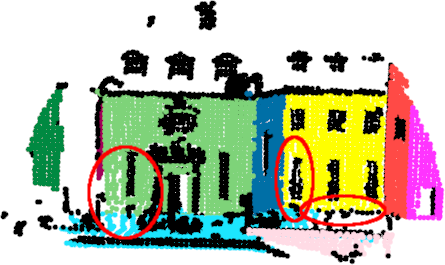}}
\caption{
Intermediate result of the relative optimization with three laser point clouds recorded on the site of Fraunhofer IAIS. The segmented surfaces are marked in color. 
The red dot indicates the respective position of the UGV. (c) shows the current point cloud, which was initially registered with point cloud (b). Point clouds 1 and 2 represent neighbouring point clouds of point cloud 3 and are summarized in (d). Point cloud 2 has gaps due to occlusions (see red markings). 
By a combination with point cloud 1, however, the gaps could be closed and thus a better calculation of the area size with respect to point cloud 3 could be carried out (see red markings in (d)).}
\label{fig:metascan}
\end{figure}

So far, the assumption has been made that the pose of the first laser point cloud is approximately known with respect to the global point cloud. In the following, an approach to determine the initial pose is presented, which only uses the structural information from the point clouds. The procedure is orientated on \cite{schmiedel2015iron} and can be described as follows:

\begin{enumerate}
\item The planes of the initial laser point clouds are segmented.
\item The global point cloud is divided into cells. The size of a cell is determined by the size of the laser point cloud plus a tolerance range. For each cell, a plane segmentation as well as a planar segment-based registration with the laser point cloud is done.
\item For each registration with a cell, the proportion of the match is calculated as follows:

\begin{equation}
\label{eq:inlier_ratio}
r = \frac{|\mathcal{K}|}{max \left( |^{m}P|, |^{d}P| \right)},
\end{equation}

where $|\mathcal{K}|$ is the number of matching planes. The respective number of segmented planes of the laser point cloud and the point cloud set by the cell is given by $|^{m}P|$ and $|^{d}P|$. The better the current cell represents the position of the laser point cloud, the greater is the number of corresponding planes. The number $|\mathcal{K }|$ of the corresponding planes therefore corresponds to a large proportion of the maximum possible number of correspondences. For cells with few common planes, the proportion of correspondences is small compared to the possible number of correspondences. The cell that best represents the position of the laser point cloud is given by the largest value $r$. If $r_1$ is the largest determined value, the following criteria must be met for a unique match:

\begin{equation}
r_1 > \alpha \quad \text{and} \quad r_1 > \beta r_2.
\end{equation}

$\alpha$ is a pre-defined threshold that $r_1$ must reach at least. $r_2$ is the second largest value given by equation~\ref{eq:inlier_ratio}. The ratio $\beta$ defines the relationship between $r_1$ and $r_2$. If these criteria cannot be fulfilled, there are several cells with a similar agreement and the best position for the laser point cloud is not clearly determinable. In this case, further laser point clouds have to be collected and re-evaluated. The global position of the laser point cloud can finally be determined by the combination of the cell position and the transformation, which was calculated in the context of the planar segment-based registration.
\end{enumerate}


\begin{table}[]
\renewcommand{\arraystretch}{1.1}
\caption{Plane segmentation of the laser point cloud.}
\label{tab:eval_rel_phoenix_time}
\centering\footnotesize
\begin{tabular}{crrrr}
\hline
Nr. & Points & Preprocessing [s] & Segmentation [s] & Planes\\
\hline
1 & 103090 & 0.0090 & 0.1989 & 8\\
2 & 135233 & 0.0096 & 0.8091 & 9\\
3 & 182623 & 0.0171 & 1.2556 & 11\\
4 & 265629 & 0.0175 & 1.4302 & 14\\
5 & 286333 & 0.0248 & 2.5137 & 19\\
6 & 291043 & 0.0197 & 3.0201 & 29\\
7 & 289934 & 0.0188 & 2.8454 & 24\\
\hline
\end{tabular}
\end{table}

\begin{table}[]
\renewcommand{\arraystretch}{1.1}
\caption{Registration of the laser point cloud.}
\label{tab:eval_rel_phoenix_log}
\centering\footnotesize
\begin{tabular}{crrr}
\hline
Pair & Registration [s] & Correspondences & ICP [s]\\
\hline
1 $\rightarrow$ 2 & 0.0116 & 2 & 0.817343\\
2 $\rightarrow$ 3 & 0.3643 & 8 & 0.231159\\
3 $\rightarrow$ 4 & 0.0814 & 7 & 0.725623\\
4 $\rightarrow$ 5 & 2.3613 & 18 & 1.02961\\
5 $\rightarrow$ 6 & 4.6153 & 15 & 0.72616\\
6 $\rightarrow$ 7 & 7.7108 & 17 & 0.839342\\
\hline
& & &\\
& & &\\
\end{tabular}
\end{table}

\begin{table}[]
\renewcommand{\arraystretch}{1.1}
\caption{Plane segmentation of the vision-based point cloud sections.}
\label{tab:eval_global_opt_phoenix_time}
\centering\footnotesize
\begin{tabular}{crrrr}
\hline
Nr. & Points & Preprocessing [s] & Segmentation [s] & Planes\\
\hline
1 & 30745 & 0.0035 & 2.4906 & 53\\
2 & 49258 & 0.0067 & 6.9224 & 109\\
3 & 53586 & 0.0079 & 11.2961 & 145\\
4 & 48600 & 0.0072 & 10.2647 & 149\\
5 & 32366 & 0.0055 & 6.2082 & 111\\
6 & 28300 & 0.0050 & 6.4789 & 108\\
7 & 9754 & 0.0028 & 2.0793 & 61\\
\hline
\end{tabular}
\end{table}

\begin{table}[]
\renewcommand{\arraystretch}{1.1}
\caption{Registration of the laser point clouds with the global vision-based point cloud.}
\label{tab:eval_global_opt_phoenix_log}
\centering\footnotesize
\begin{tabular}{crrr}
\hline
Pair & Registration [s] & Correspondences & ICP [s]\\
\hline
1 Laser $\rightarrow$ Camera & 0.012568 & 6 & -\\
2 Laser $\rightarrow$ Camera & 0.2772 & 7 & 0.480169\\
3 Laser $\rightarrow$ Camera & 0.5608 & 11 & -\\
4 Laser $\rightarrow$ Camera & 0.4258 & 9 & 0.630182\\
5 Laser $\rightarrow$ Camera & 0.5098 & 14 & -\\
6 Laser $\rightarrow$ Camera & 1.3474 & 20 & -\\
7 Laser $\rightarrow$ Camera & 0.0896 & 0 & 0.861286\\
\hline
\end{tabular}
\end{table}

\begin{table}[]
\renewcommand{\arraystretch}{1.1}
\caption{Relative pose error and absolute trajectory error of the relative localization.}
\label{tab:eval_rel_phoenix_rpe_ate}
\centering\footnotesize
\begin{tabular}{lrrr}
\hline
Metric & $E_{rmse}$ [m] & $E_{min}$ [m] & $E_{max}$ [m]\\
\hline
RPE & 3.3579 & 0.2090 & 8.1820\\
ATE & 3.1074 & 0.6780 & 7.4754\\
\hline
& & &\\
& & &\\
& & &\\
& & &\\
\end{tabular}
\end{table}

\section{Experiments}
\label{sec:experiments}
In this section the results of the planar segment-based localization are evaluated.
For the test environment the former site of the blast furnace Phoenix-West in Dortmund was selected (see \cref{fig:phoenix_west}).

The UGV started near the entrance area of the factory building, drove further into the hall and finished the recordings there. A total of 7 laser scans were recorded. The relative pose error and absolute trajectory error (RPE / ATE) of the estimated trajectory after \cite{sturm12iros} were used as evaluation criteria for the localization. In order to obtain a reference trajectory of the UGV, adjacent laser point clouds were aligned relative to each other and the poses were subsequently refined with the SLAM framework 3DTK \cite{nuchter20076d}.
The vision-based point cloud for the following experiments was generated by MVE with images at a resolution of $640 \times 480$ pixels. MVE was choosen since it provides convincing results with respect to the estimated trajectory as well as the Mean Plane Variance (MPV) and Mean Map Entropy (MME), which were computed according to \cite{razlaw2015evaluation}. However the approach presented in this work is not limited to MVE.

The processing times of the pose tracking without further optimization steps are listed in the tables \ref{tab:eval_rel_phoenix_time} and \ref{tab:eval_rel_phoenix_log}. The RPE and ATE are represented in \cref{tab:eval_rel_phoenix_rpe_ate} and \cref{fig:eval_rel_phoenix_results_rpe_ate}. The errors in the trajectory are caused by less accurate registration of the first two point clouds. The reason for this are inadequate structural elements, that do not allow accurate estimation of all directions of translation (see \cref{fig:eval_rel_phoenix_cor}). For the globally optimized trajectory, the results listed in \cref{tab:eval_global_opt_phoenix_time} and \cref{tab:eval_global_opt_phoenix_log} were obtained. The deviations in the trajectory could be reduced by the optimization. In contrast to the relative localization, the global point cloud enabled the correct registration of the first two point clouds by additional structures (see \cref{fig:eval_global_opt_phoenix_cor} and \cref{fig:eval_global_opt_phoenix_results_rpe_ate}, for the RPE and ATE). The overall registration result is shown in \cref{fig:eval_global_opt_phoenix_results_complete}. The initial localization was evaluated as a final test. The test sequence locates each laser point cloud in the global point cloud. The first five point clouds could be located correctly. The two last point clouds represented areas within the factory and had too little overlap with the global point cloud.\\

\begin{figure}[]
\centering
\subfloat[RPE]{\includegraphics[width=0.49\linewidth]{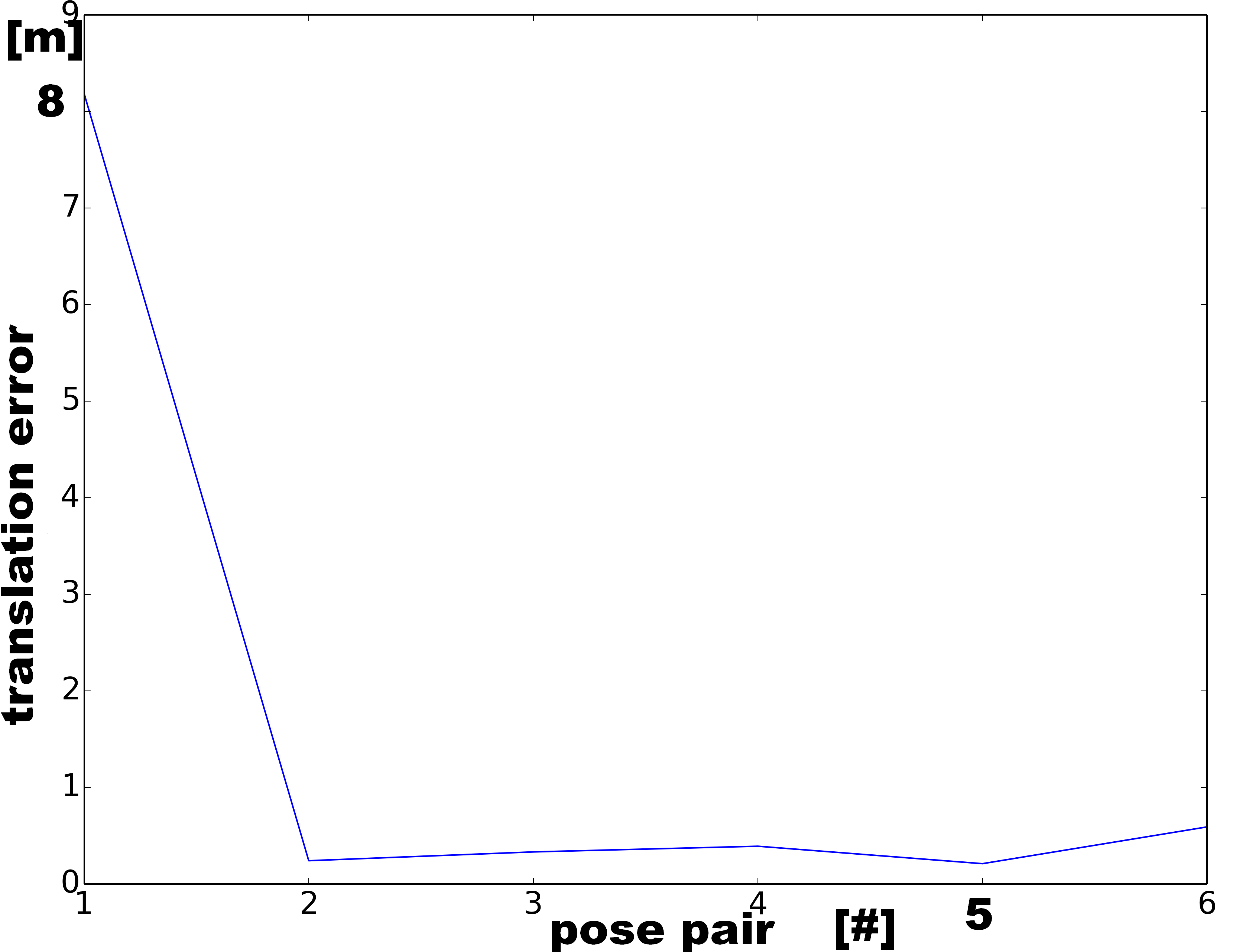}}
\hfil
\subfloat[ATE]{\includegraphics[width=0.49\linewidth]{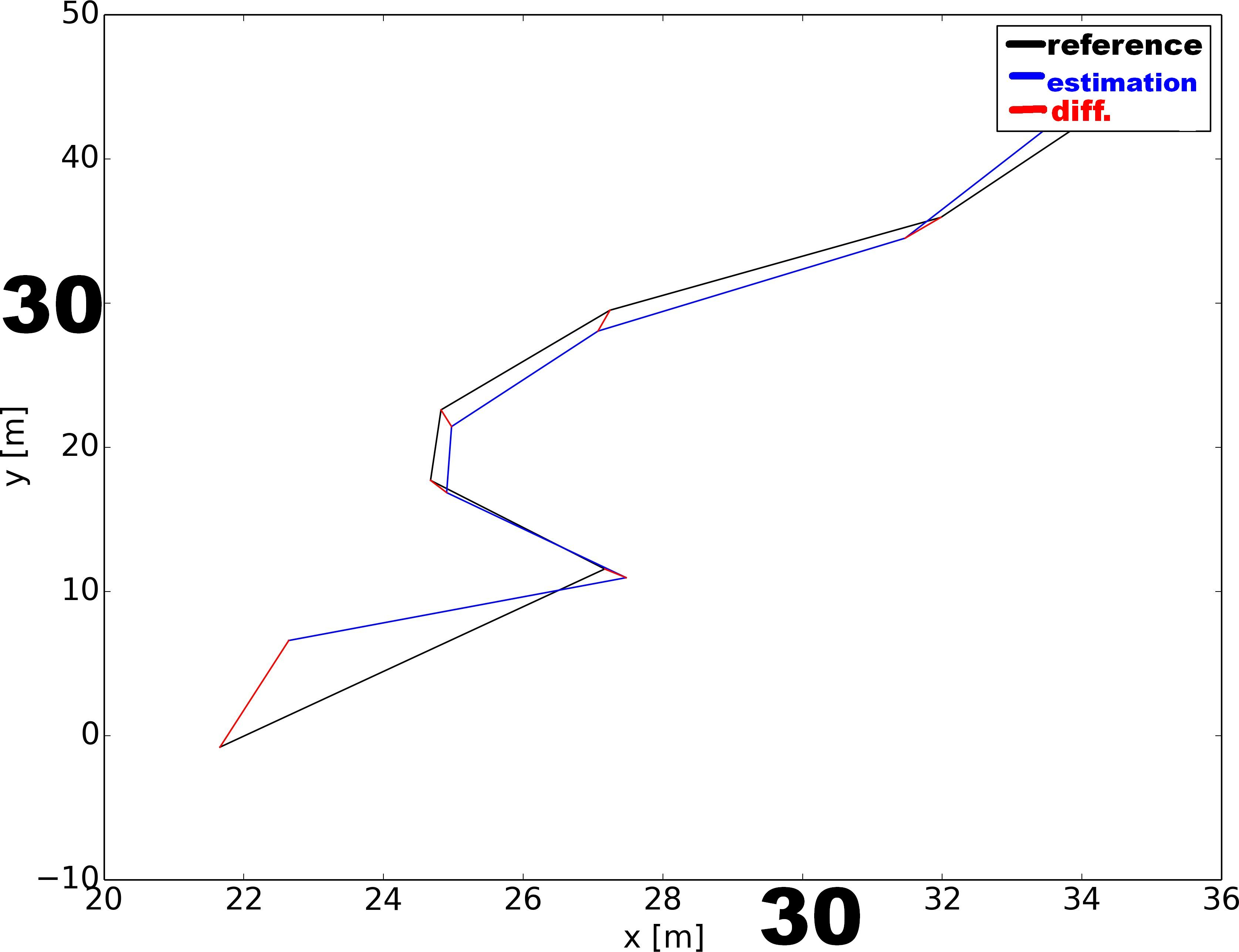}}
\caption{RPE and ATE of the relative localization.}
\label{fig:eval_rel_phoenix_results_rpe_ate}
\end{figure}

\begin{figure}[]
\centering
\subfloat[RPE]{\includegraphics[width=0.49\linewidth]{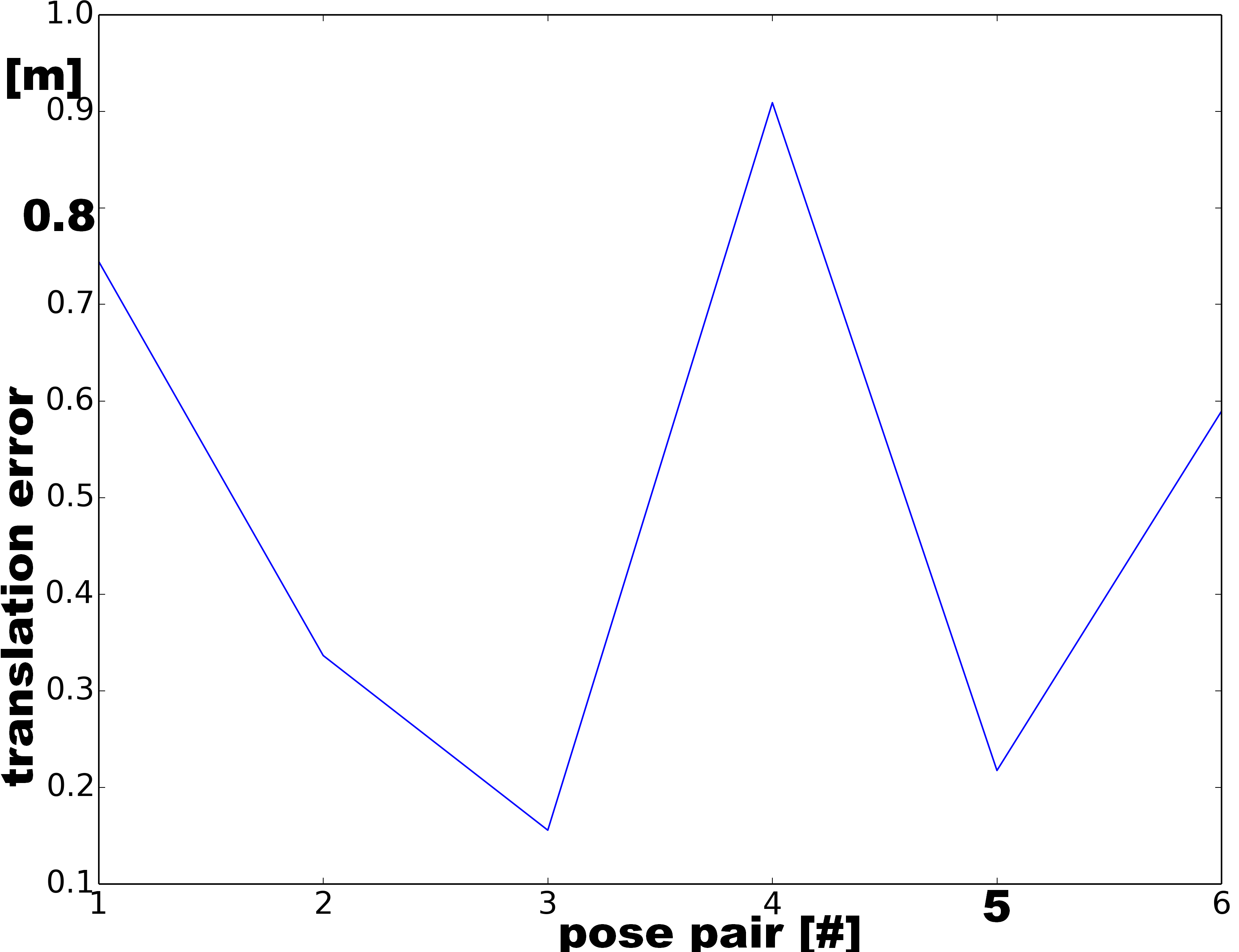}}
\hfil
\subfloat[ATE]{\includegraphics[width=0.49\linewidth]{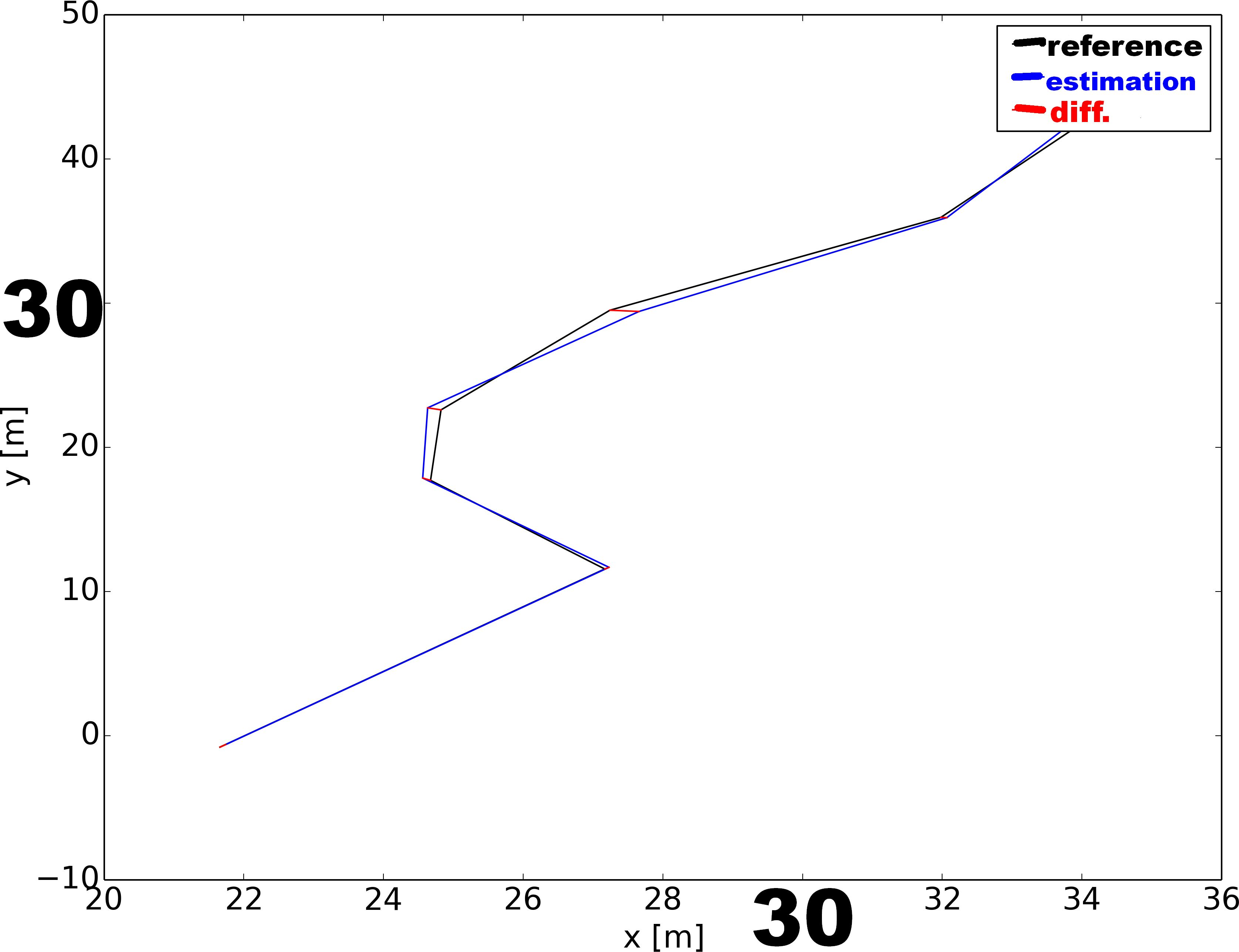}}
\caption{RPE and ATE of the globally optimized localization.}
\label{fig:eval_global_opt_phoenix_results_rpe_ate}
\end{figure}

\begin{figure}[]
\centering
\subfloat[From: Point cloud 1 to]{\includegraphics[width=0.49\linewidth]{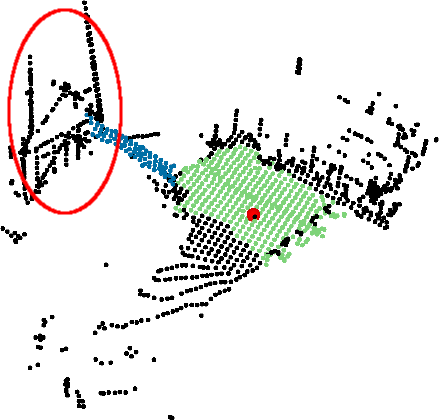}}
\hfil
\subfloat[Point cloud 2]{\includegraphics[width=0.49\linewidth]{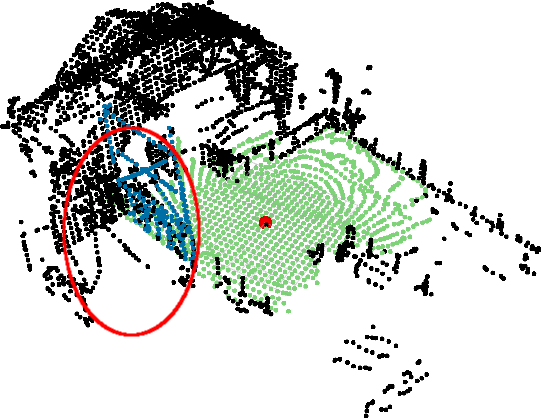}}
\hfil
\subfloat[From: Point cloud 2 to]{\includegraphics[width=0.49\linewidth]{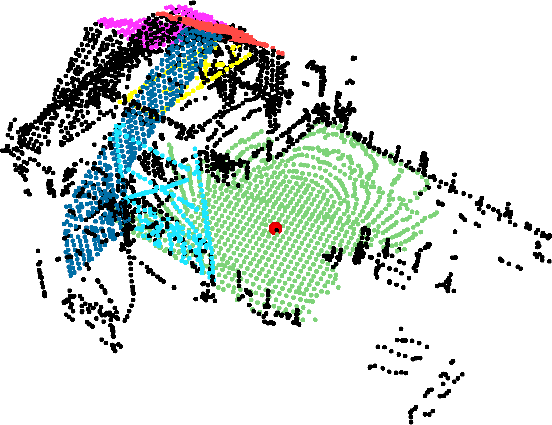}}
\hfil
\subfloat[Point cloud 3]{\includegraphics[width=0.49\linewidth]{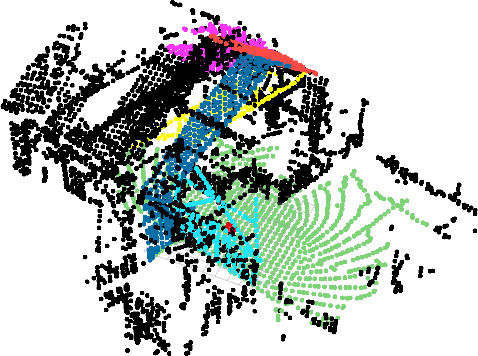}}
\caption{Registration of the first three point clouds. Corresponding planes are randomly colored. The red marked area in image a could not be captured from the laser scanner in image b and thus offers no possibility
for determining the direction of translation.}
\label{fig:eval_rel_phoenix_cor}
\end{figure}

\begin{figure}[]
\centering
\subfloat[Point cloud 1]{\includegraphics[width=0.49\linewidth]{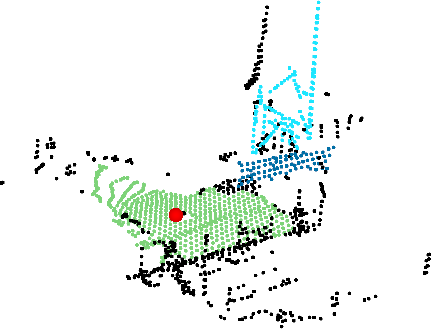}}
\hfil
\subfloat[Section 1]{\includegraphics[width=0.49\linewidth]{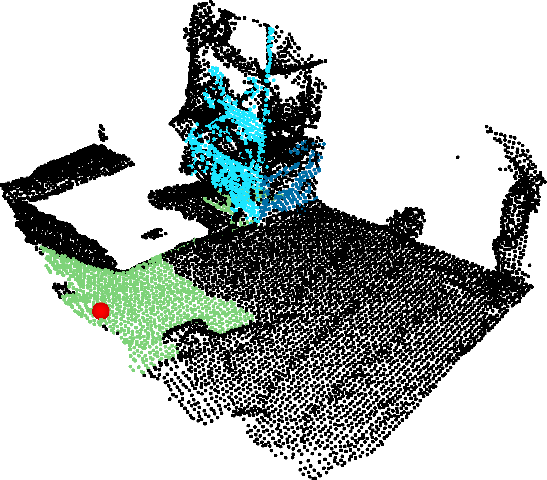}}
\hfil
\subfloat[Point cloud 2]{\includegraphics[width=0.49\linewidth]{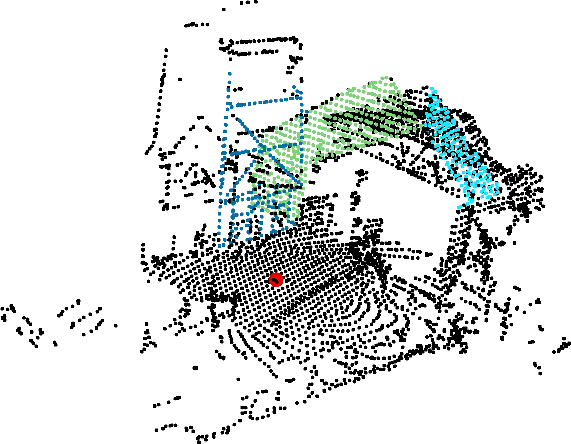}}
\hfil
\subfloat[Section 2]{\includegraphics[width=0.49\linewidth]{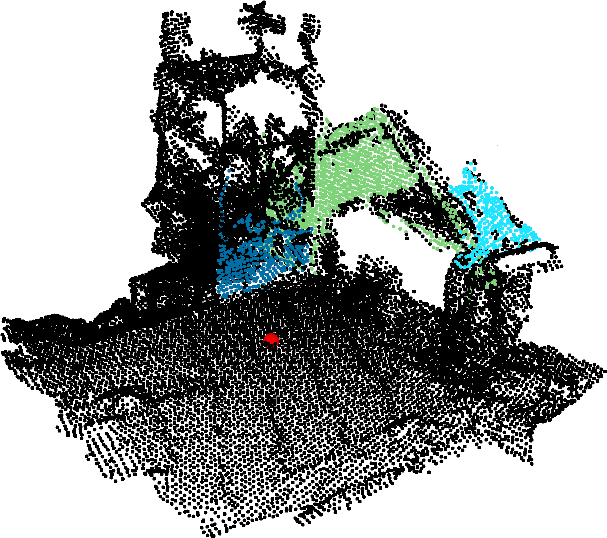}}
\caption{Registered pairs of the globally optimized localization.}
\label{fig:eval_global_opt_phoenix_cor}
\end{figure}

\begin{figure*}[]
\centering
\includegraphics[width=0.99\textwidth]{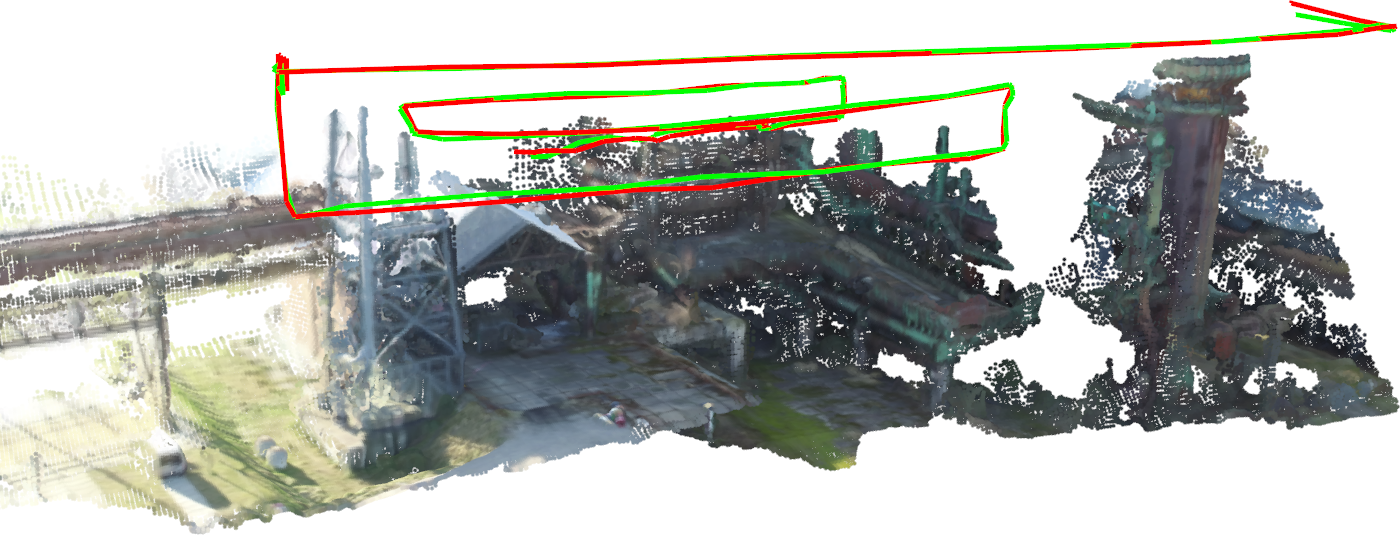}
\caption{Merged point cloud of the former site of the blast furnace Phoenix-West in Dortmund. The green trajectory was determined with MVE and the red trajectory with GPS. Both trajectories overlap to a large amount, which shows a good estimation of the trajectory. The point clouds were generated with MVE at a resolution of $640 \times 480$ pixels.}
\label{fig:phoenix_west}
\end{figure*}

\begin{figure*}[]
\centering
\includegraphics[width=0.99\textwidth]{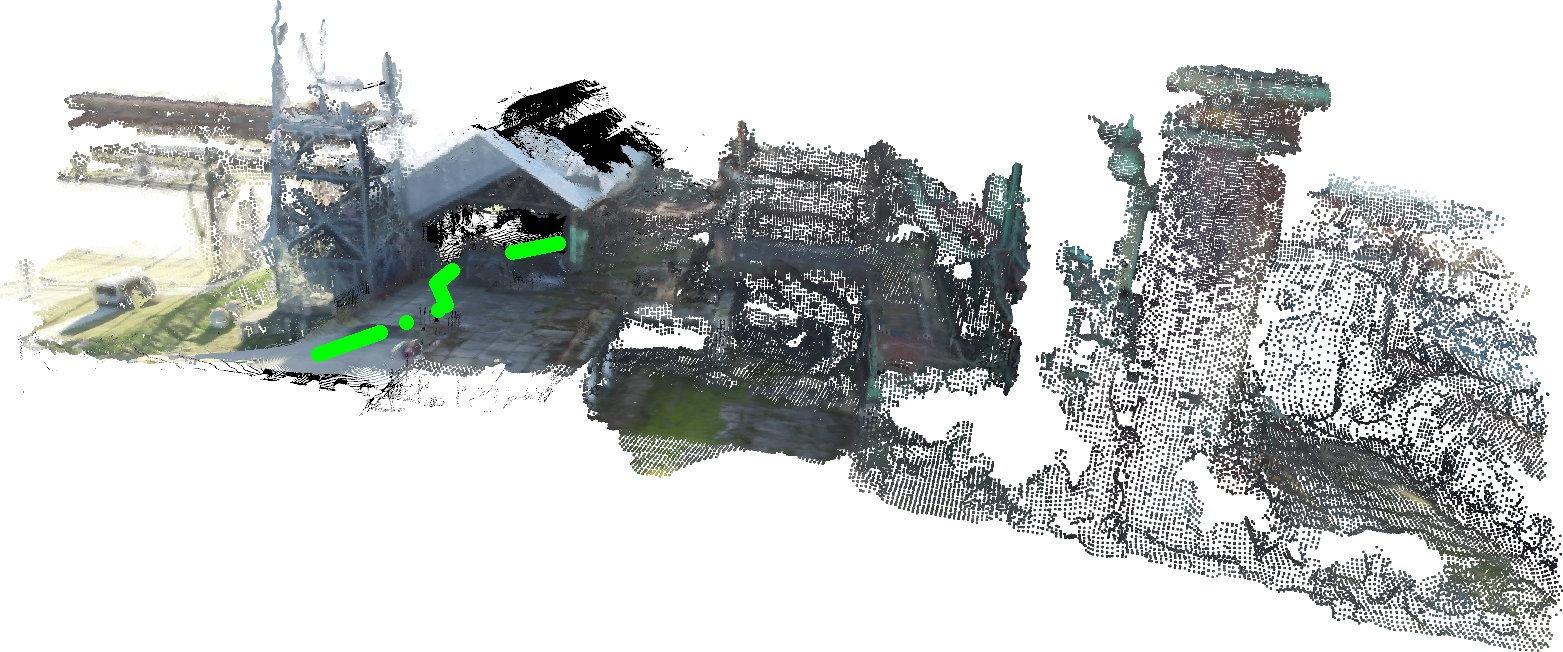}
\caption{Registered point clouds of the globally optimized localization. Points of the laser point clouds were dyed with the color information of the global point cloud. Within the factory no color information could be extracted. The estimated trajectory is shown in green.}
\label{fig:eval_global_opt_phoenix_results_complete}
\end{figure*}

\section{Conclusion and future work}
The base for human to robot and robot to robot collaboration is a persistent environment model, which implies to fuse different sensor streams of different modalities.
We present a novel approach for the plane-based localization of laser point clouds (UGV) in monocular vision point clouds (UAV).
The method first performs a plane segmentation and then attempts to register neighbouring point clouds by means of corresponding planes.
The method uses a global point cloud generated by the UAV's camera recordings for optimization.
The evaluation showed that the relative localization provided a reliable registration and is therefore suitable as an initial estimation for global optimization.
Point clouds, which had inadequate structures or slight overlaps with neighbouring point clouds, prevented accurate registration.
Differences in the relative localization could be offset by the global optimization.
A further important component of the global localization is the determination of the initial pose of the UGV.
We suggested an automatic search of the start sector with subsequent registration.
Areas with more than 50\% overlap were successfully localized.
When evaluating the vision-based procedures, it turned out that MVE (\cite{fuhrmann2014mve}) is best suited for planar segment-based registration.

The localization method developed in this work will be extended for future work, e.g. by the utilization of additional sensor information.
For example, laser point clouds with color information can be generated by the camera on the UGV.
For the correspondence search, this color information, in addition to the surface area, forms a further useful criterion for matching surfaces.
When possible, GPS coordinates can also be used to support the localization.


\bibliographystyle{IEEEtran}
\bibliography{IEEEabrv,literatur}

\begin{thebibliography}{10}
\providecommand{\url}[1]{#1}
\csname url@rmstyle\endcsname
\providecommand{\newblock}{\relax}
\providecommand{\bibinfo}[2]{#2}
\providecommand\BIBentrySTDinterwordspacing{\spaceskip=0pt\relax}
\providecommand\BIBentryALTinterwordstretchfactor{4}
\providecommand\BIBentryALTinterwordspacing{\spaceskip=\fontdimen2\font plus
\BIBentryALTinterwordstretchfactor\fontdimen3\font minus
  \fontdimen4\font\relax}
\providecommand\BIBforeignlanguage[2]{{%
\expandafter\ifx\csname l@#1\endcsname\relax
\typeout{** WARNING: IEEEtran.bst: No hyphenation pattern has been}%
\typeout{** loaded for the language `#1'. Using the pattern for}%
\typeout{** the default language instead.}%
\else
\language=\csname l@#1\endcsname
\fi
#2}}

\bibitem{murphy2000marsupial}
R.~R. Murphy, ``Marsupial and shape-shifting robots for urban search and
  rescue,'' \emph{Intelligent Systems and their Applications, IEEE}, vol.~15,
  no.~2, pp. 14--19, 2000.

\bibitem{holz2015registration}
D.~Holz, A.~E. Ichim, F.~Tombari, R.~B. Rusu, and S.~Behnke, ``Registration
  with the point cloud library: A modular framework for aligning in 3-{D},''
  \emph{IEEE Robotics \& Automation Magazine}, vol.~22, no.~4, pp. 110--124,
  2015.

\bibitem{klein07parallel}
G.~Klein and D.~Murray, ``Parallel tracking and mapping for small {AR}
  workspaces,'' in \emph{Proc. Sixth {IEEE} and {ACM} International Symposium
  on Mixed and Augmented Reality {(ISMAR'07)}}, Nara, Japan, November 2007.

\bibitem{davison2007monoslam}
A.~J. Davison, I.~D. Reid, N.~D. Molton, and O.~Stasse, ``Monoslam: Real-time
  single camera {SLAM},'' \emph{IEEE transactions on pattern analysis and
  machine intelligence}, vol.~29, no.~6, pp. 1052--1067, 2007.

\bibitem{newcombe2011dtam}
R.~A. Newcombe, S.~J. Lovegrove, and A.~J. Davison, ``{DTAM}: Dense tracking
  and mapping in real-time,'' in \emph{2011 International Conference on
  Computer Vision}.\hskip 1em plus 0.5em minus 0.4em\relax IEEE, 2011, pp.
  2320--2327.

\bibitem{engel2013iccv}
J.~Engel, J.~Sturm, and D.~Cremers, ``Semi-dense visual odometry for a
  monocular camera,'' in \emph{IEEE International Conference on Computer Vision
  (ICCV)}, Sydney, Australia, December 2013.

\bibitem{Forster2014ICRA}
C.~Forster, M.~Pizzoli, and D.~Scaramuzza, ``{SVO}: Fast semi-direct monocular
  visual odometry,'' in \emph{IEEE International Conference on Robotics and
  Automation (ICRA)}, 2014.

\bibitem{engel14eccv}
J.~Engel, T.~Sch\"ops, and D.~Cremers, ``{LSD-SLAM}: Large-scale direct
  monocular {SLAM},'' in \emph{European Conference on Computer Vision (ECCV)},
  September 2014.

\bibitem{mur2015probabilistic}
R.~Mur-Artal and J.~D. Tard{\'o}s, ``Probabilistic semi-dense mapping from
  highly accurate feature-based monocular {SLAM},'' \emph{Proceedings of
  Robotics: Science and Systems, Rome, Italy}, vol.~1, 2015.

\bibitem{murAcceptedTRO2015}
R.~Mur-Artal, J.~M.~M. Montiel, and J.~D. Tard\'os, ``{ORB-SLAM}: a versatile
  and accurate monocular {SLAM} system,'' \emph{IEEE Transactions on Robotics},
  vol.~31, no.~5, pp. 1147--1163, 2015.

\bibitem{stuehmer-et-al-dagm10}
J.~St\"uhmer, S.~Gumhold, and D.~Cremers, ``Real-time dense geometry from a
  handheld camera,'' in \emph{Pattern Recognition (Proc. DAGM)}, Darmstadt,
  Germany, September 2010, pp. 11--20.

\bibitem{Pizzoli2014ICRA}
M.~Pizzoli, C.~Forster, and D.~Scaramuzza, ``{REMODE}: Probabilistic, monocular
  dense reconstruction in real time,'' in \emph{IEEE International Conference
  on Robotics and Automation (ICRA)}, 2014.

\bibitem{conchaIROS15}
A.~Concha and J.~Civera, ``{Dense Piecewise Planar Tracking and Mapping from a
  Monocular Sequence},'' in \emph{Proc. of The International Conference on
  Intelligent Robots and Systems (IROS)}, 2015.

\bibitem{felzenszwalb2004efficient}
P.~F. Felzenszwalb and D.~P. Huttenlocher, ``Efficient graph-based image
  segmentation,'' \emph{International Journal of Computer Vision}, vol.~59,
  no.~2, pp. 167--181, 2004.

\bibitem{fuhrmann2014mve}
S.~Fuhrmann, F.~Langguth, and M.~Goesele, ``{MVE} -- a multiview reconstruction
  environment,'' in \emph{Proceedings of the Eurographics Workshop on Graphics
  and Cultural Heritage (GCH)}, vol.~6, no.~7, 2014, p.~8.

\bibitem{Pathak2009}
K.~Pathak, N.~Vaskevicius, J.~Poppinga, M.~Pfingsthorn, S.~Schwertfeger, and
  A.~Birk, ``Fast {3D} mapping by matching planes extracted from range sensor
  point-clouds,'' in \emph{Intelligent Robots and Systems, 2009. IROS 2009.
  IEEE/RSJ International Conference on}, 2009.

\bibitem{Besl1992}
P.~J. Besl and N.~D. McKay, ``A method for registration of {3-D} shapes,''
  \emph{IEEE Transactions on Pattern Analysis and Machine Intelligence},
  vol.~14, no.~2, pp. 239--256, 1992.

\bibitem{viejo20073d}
D.~Viejo and M.~Cazorla, ``{3D} plane-based egomotion for {SLAM} on
  semi-structured environment,'' in \emph{2007 IEEE/RSJ International
  Conference on Intelligent Robots and Systems}.\hskip 1em plus 0.5em minus
  0.4em\relax IEEE, 2007, pp. 2761--2766.

\bibitem{PathakJFR2010}
K.~Pathak, A.~Birk, N.~Vaskevicius, M.~Pfingsthorn, S.~Schwertfeger, and
  J.~Poppinga, ``Online three-dimensional {SLAM} by registration of large
  planar surface segments and closed-form pose-graph relaxation,''
  \emph{Journal of Field Robotics}, vol.~27, no.~1, pp. 52--84, 2010.

\bibitem{poppinga2008fast}
J.~Poppinga, N.~Vaskevicius, A.~Birk, and K.~Pathak, ``Fast plane detection and
  polygonalization in noisy {3D} range images,'' in \emph{2008 IEEE/RSJ
  International Conference on Intelligent Robots and Systems}.\hskip 1em plus
  0.5em minus 0.4em\relax IEEE, 2008, pp. 3378--3383.

\bibitem{hahnel2003learning}
D.~H{\"a}hnel, W.~Burgard, and S.~Thrun, ``Learning compact {3D} models of
  indoor and outdoor environments with a mobile robot,'' \emph{Robotics and
  Autonomous Systems}, vol.~44, no.~1, pp. 15--27, 2003.

\bibitem{pathak2010fast}
K.~Pathak, A.~Birk, N.~Vaskevicius, and J.~Poppinga, ``Fast registration based
  on noisy planes with unknown correspondences for {3-D} mapping,'' \emph{IEEE
  Transactions on Robotics}, vol.~26, no.~3, pp. 424--441, 2010.

\bibitem{xiao2013planar}
J.~Xiao, B.~Adler, J.~Zhang, and H.~Zhang, ``Planar segment based
  three-dimensional point cloud registration in outdoor environments,''
  \emph{Journal of Field Robotics}, vol.~30, no.~4, pp. 552--582, 2013.

\bibitem{gawel2016structure}
A.~Gawel, T.~Cieslewski, R.~Dub{\'e}, M.~Bosse, R.~Siegwart, and J.~Nieto,
  ``Structure-based vision-laser matching,'' in \emph{Intelligent Robots and
  Systems (IROS), 2016 IEEE/RSJ International Conference on}.\hskip 1em plus
  0.5em minus 0.4em\relax IEEE, 2016, pp. 182--188.

\bibitem{Marton09ICRA}
Z.~C. Marton, R.~B. Rusu, and M.~Beetz, ``{On Fast Surface Reconstruction
  Methods for Large and Noisy Datasets},'' in \emph{Proceedings of the IEEE
  International Conference on Robotics and Automation (ICRA)}, Kobe, Japan, May
  12-17 2009.

\bibitem{surmann2003autonomous}
H.~Surmann, A.~N{\"u}chter, and J.~Hertzberg, ``An autonomous mobile robot with
  a {3D} laser range finder for {3D} exploration and digitalization of indoor
  environments,'' \emph{Robotics and Autonomous Systems}, vol.~45, no.~3, pp.
  181--198, 2003.

\bibitem{schmiedel2015iron}
T.~Schmiedel, E.~Einhorn, and H.-M. Gross, ``{IRON: A fast interest point
  descriptor for robust NDT-map matching and its application to robot
  localization},'' in \emph{Intelligent Robots and Systems (IROS), 2015
  IEEE/RSJ International Conference on}.\hskip 1em plus 0.5em minus 0.4em\relax
  IEEE, 2015, pp. 3144--3151.

\bibitem{sturm12iros}
J.~Sturm, N.~Engelhard, F.~Endres, W.~Burgard, and D.~Cremers, ``A benchmark
  for the evaluation of {RGB-D SLAM} systems,'' in \emph{Proc. of the
  International Conference on Intelligent Robot Systems (IROS)}, Oct. 2012.

\bibitem{nuchter20076d}
A.~N{\"u}chter, K.~Lingemann, J.~Hertzberg, and H.~Surmann, ``{6D SLAM} -- {3D}
  mapping outdoor environments,'' \emph{Journal of Field Robotics (JFR)},
  vol.~24, no. 8-9, pp. 699--722, 2007.

\bibitem{razlaw2015evaluation}
J.~Razlaw, D.~Droeschel, D.~Holz, and S.~Behnke, ``Evaluation of registration
  methods for sparse {3D} laser scans,'' in \emph{Mobile Robots (ECMR), 2015
  European Conference on}.\hskip 1em plus 0.5em minus 0.4em\relax IEEE, 2015,
  pp. 1--7.

\end{thebibliography}

\end{document}